%% file: main.tex
\documentclass[10pt,twocolumn,letterpaper]{article}

\usepackage{iccv}              
\usepackage{svg}
\usepackage{adjustbox}
\usepackage{multirow}


\definecolor{iccvblue}{rgb}{0.21,0.49,0.74}
\usepackage[pagebackref,breaklinks,colorlinks,allcolors=iccvblue]{hyperref}

\def\paperID{} 
\def\confName{ArXiv}
\def\confYear{2025}

\title{Revisiting Deepfake Detection: Chronological Continual Learning and the Limits of Generalization}
\author{Federico Fontana\\
Sapienza University of Rome\\
{\tt\small federico.fontana@uniroma1.it}
\and
Anxhelo Diko\\
Sapienza University of Rome\\
{\tt\small anxhelo.diko@uniroma1.it}
\and
Romeo Lanzino\\
Sapienza University of Rome\\
{\tt\small romeo.lanzino@uniroma1.it}
\and
Marco Raoul Marini\\
Sapienza University of Rome\\
{\tt\small marco.marini@uniroma1.it}
\and
Bachir Kaddar\\
University Ibn Khaldoun of Tiaret\\
{\tt\small b.kaddar@univ-tiaret.dz}
\and
Gian Luca Foresti\\
University of Udine\\
{\tt\small gianluca.foresti@uniud.it}
\and
Luigi Cinque\\
Sapienza University of Rome\\
{\tt\small luigi.cinque@uniroma1.it}
}

\begin{document}
\maketitle
\input{sec/0_abstract}    
\input{sec/1_intro}
\input{sec/2_related_work}
\input{sec/4_method}

\input{sec/5_experiments}

\input{sec/6_ablation_study}

\input{sec/7_conclusion}

{
    \small
    \bibliographystyle{ieeenat_fullname}
    \bibliography{main}
}

\end{document}

%% file: sec/0_abstract.tex
\begin{abstract}
The rapid evolution of deepfake generation technologies poses critical challenges for detection systems, as non-continual learning methods demand frequent and expensive retraining. We reframe deepfake detection (DFD) as a Continual Learning (CL) problem, proposing an efficient framework that incrementally adapts to emerging visual manipulation techniques while retaining knowledge of past generators. Our framework, unlike prior approaches that rely on unreal simulation sequences, simulates the real-world chronological evolution of deepfake technologies in extended periods across 7 years. Simultaneously, our framework builds upon lightweight visual backbones to allow for the real-time performance of DFD systems.  Additionally, we contribute two novel metrics: Continual AUC (C-AUC) for historical performance and Forward Transfer AUC (FWT-AUC) for future generalization. Through extensive experimentation (over 600 simulations), we empirically demonstrate that while efficient adaptation (+155 times faster than full retraining) and robust retention of historical knowledge is possible, the generalization of current approaches to future generators without additional training remains near-random (FWT-AUC $\approx$ 0.5) due to the unique imprint characterizing each existing generator. Such observations are the foundation of our newly proposed Non-Universal Deepfake Distribution Hypothesis. 
 \textbf{Code will be released upon acceptance.}
\end{abstract}


%% file: sec/1_intro.tex
\section{Introduction}
\label{sec:intro}
\begin{figure}[ht]
    \centering
    \includegraphics[width=\linewidth]{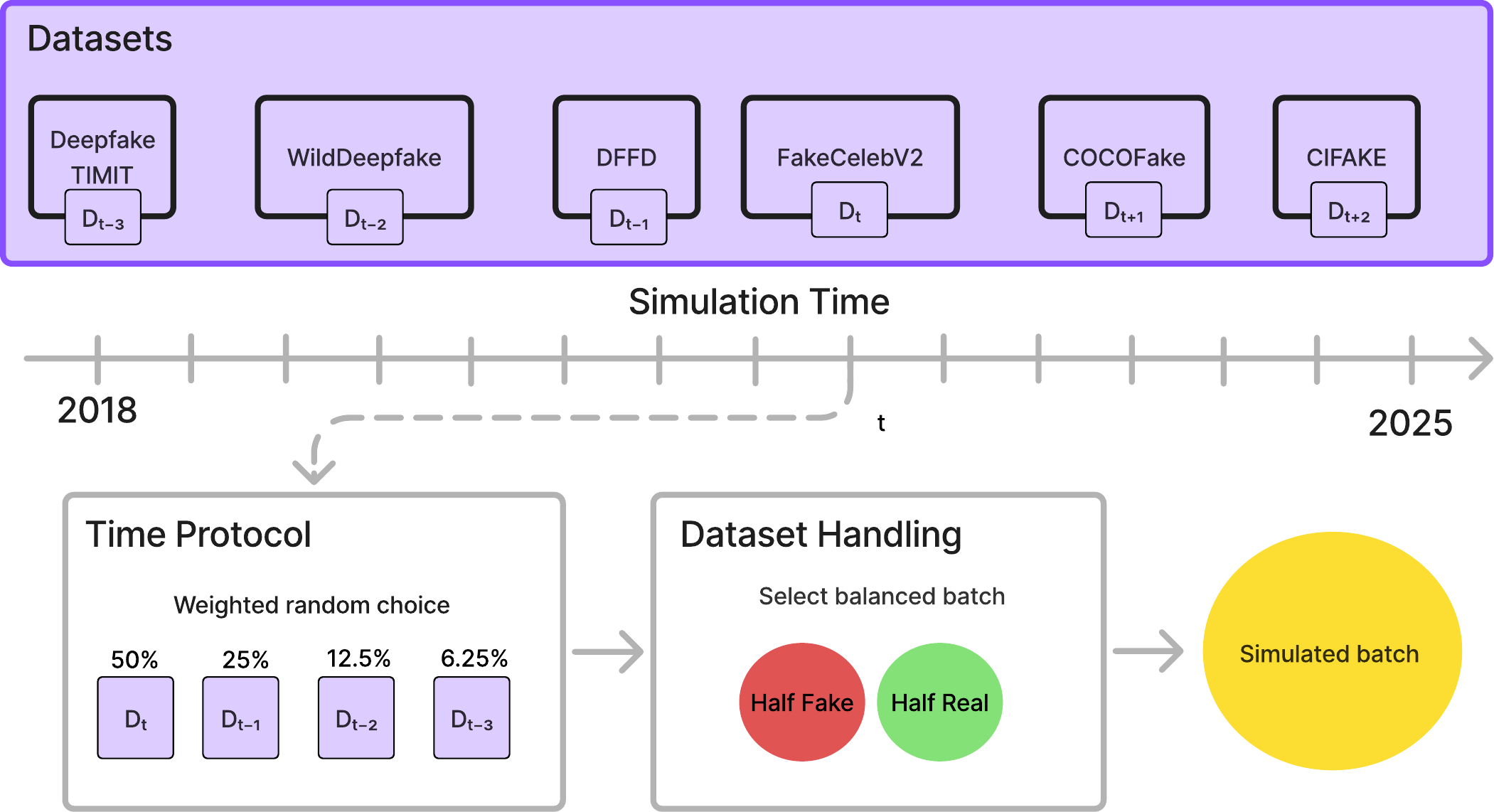}
    \caption{Illustration of the simulation framework spanning from 2018 to 2025. Given a time \( t \), a dataset is selected according to a Time protocol followed by a custom batch extraction which simulates a real-world data stream scenario at the specified time \( t \).}
    \label{fig:framework}
\vspace{-5mm}
\end{figure}
The evolution of deepfake technology has profound implications for the authenticity of digital media. With novel deepfake generators emerging continuously, traditional deepfake detection (DFD) methods—which rely on static training paradigms—struggle to keep pace. This leads not only to significant computational overhead but also to reduced effectiveness in dynamic, real-world environments. Such challenges are especially critical on social media, where the rapid spread of manipulated content can exacerbate misinformation.
Prior works \cite{li2023continual,ieee10034569, TASSONE2024104143} have explored the application of continual learning (CL) for DFD. \cite{li2023continual} introduces a continual learning benchmark for DFD with three evaluation sequences based on difficulty or length, applying CL methods in class-incremental (CIL) and domain-incremental (DIL) settings. \cite{TASSONE2024104143} builds on this by adding a new generator and evaluating two CL methods, while \cite{ieee10034569} reframes DFD as a CIL, focusing on DFD generalization. 

However, current approaches prove useful only in a small number of CL methods \cite{TASSONE2024104143}, utilize a very limited evaluation set \cite{ieee10034569} ($\leq$2K samples), or employ a not chronological sequence of deepfake generators \cite{li2023continual,ieee10034569, TASSONE2024104143}. To our knowledge, no prior work addresses the DFD as a real-world scenario; rather, the sequence of generators fed to the CL methods is chosen either randomly, based on generator affinity, or according to the difficulty detectors face in recognizing them.
As a result, existing research cannot be directly related to real-world settings, as the experimental assumptions for training and evaluation do not reflect real conditions.
Furthermore, no in-depth study on data efficiency has been conducted for CL methods, nor has a study on training and inference efficiency been made. Without efficient training and inference, detection models risk becoming obsolete and unable to keep pace with the continuous advancements. Optimizing efficiency ensures maintainability, making DFD systems viable for deployment in high-stakes environments such as social media monitoring.

Motivated by these works, we propose a pipeline that chronologically simulates a deepfake scenario over extended periods using datasets ordered by time, along with a strategy to simulate the sequence of tasks as a random sample from a datastream in any moment(e.g., social media) as illustrated in Fig.~\ref{fig:framework}. This design aims to better capture a real-world deepfake scenario for training and evaluation while avoiding the choice bias on the generator sequence. Additionally, we frame the problem as a DIL task to maintain a detector with consistent size, speed, and memory usage throughout its lifetime. Our focus also extends to the need for data, training, and inference efficiency. Therefore, the proposed method employs lightweight models that can achieve real-time performance on most smartphones and requires significantly lower data to be trained due to the CL strategies, while maintaining competitive performance. 

We conducted a total of over 600 full simulations on the presented chronological framework and observed a striking pattern: while some methods successfully retained knowledge of past deepfake generators, none exhibited strong generalization to future ones. To systematically capture this phenomenon, we introduce two new evaluation metrics to measure the retention of past knowledge in highly imbalanced classes and one to quantify a model's generalizability to chronologically unseen deepfake generators on unbalanced classes.
Our analysis using these metrics revealed a fundamental decorrelation between past and future generalization, challenging the assumption that deepfake detection models can generalize across time. This leads us to formally propose the Non-Universal Deepfake Distribution Hypothesis, which states that deepfake detection cannot be effectively generalized through static training because each deepfake generator imprints a unique, non-transferable signature. We empirically validate this hypothesis. \\
In summary, our main contributions are:\begin{itemize}
\item We propose a novel \textbf{Chronological CL framework} for DFD that mirrors real-world deepfake evolution.
\item We evaluated \textbf{8 different CL strategies and 4 models} within the proposed framework, demonstrating adaptation while efficiently preserving historical knowledge.  
    \item We introduce \textbf{Continual AUC (C-AUC)} and \textbf{Forward Transfer AUC (FWT-AUC)} metrics to measure the stability and transferability of the learned detection capabilities in a chronological and unbalanced classes setting.
\item We propose the \emph{Non-Universal Deepfake Distribution Hypothesis}, suggesting that each deepfake generator leaves a unique, non-transferable signature, requiring continuous model updates. We support this hypothesis with empirical evaluation on more than 600 simulations.  

\end{itemize}

   

%% file: sec/2_related_work.tex
\section{Related Work}
\label{sec:related}
\begin{figure*}[htbp]
  \centering
  \includegraphics[width=\textwidth]{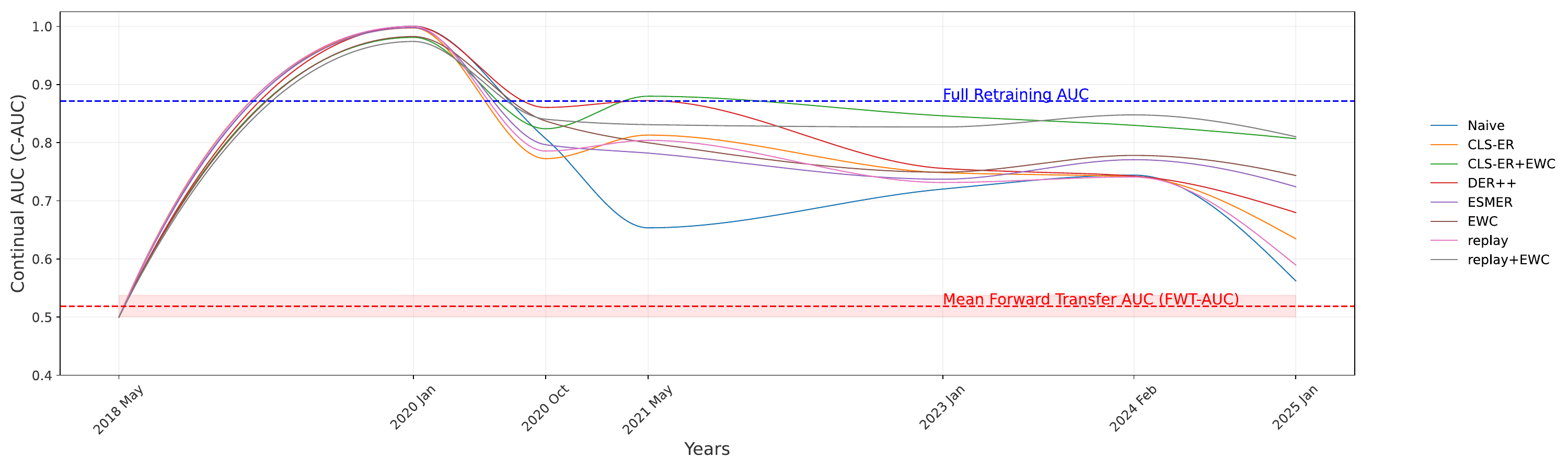}
 \caption{The lines represent the 8 CL methods evaluated on MobileNetV4, each configured with its optimal hyperparameters found in the ablation study. The C-AUC (Eq. \ref{eq_cauc}) indicates the classifiers' performance on the test data available up to that point. The red horizontal line represents the mean FWT-AUC (Eq. \ref{eq_fwtauc}) across all experiments, reflecting the methods' generalization performance on unseen datasets, with the red bars indicating the standard deviation. The blue horizontal line denotes the Eval AUC achieved by the full-retraining method.}
  \label{fig:double}
\end{figure*}
\subsection{Continual Learning}
CL tackles the challenge of catastrophic forgetting \cite{catastrophic_forgetting}, where models lose previously acquired knowledge when sequentially trained on new tasks.
In DIL\cite{van2019three}, tasks share the same label space but differ in input distributions, and task identities are not provided during training or testing.
 A naive sequential training method, lacking any CL mechanism, serves as a baseline for comparison. More advanced approaches include the Replay strategy \cite{rolnick2019experience}, which retains and replays stored samples, and Elastic Weight Consolidation (EWC) \cite{kirkpatrick2017overcoming}, a regularization method that penalizes significant alterations to the weights critical for past tasks. Based on these, Class-Semantic Replay (CLS-ER) \cite{arani2022learning} leverages semantic information to focus memory retention on task-relevant features, while Error-Sensitive Memory Replay (ESMER) \cite{sarfrazerror} dynamically adjusts the contribution of stored samples based on their error sensitivity. Additionally, Dark Experience Replay++ (DER++) \cite{buzzega2020dark} incorporates knowledge distillation to transfer useful information between tasks,  improving performance in CL scenarios.
 \subsection{Deepfake Detection}
Recent advances in deepfake detection have shifted from exploiting low-level artifacts—such as unnatural blinking \cite{li2018ictu} and inconsistent physiological signals \cite{ciftci2020fakecatcher}—to focusing on facial warping \cite{li2020celeb}, head pose anomalies \cite{yang2019exposing}, and textural irregularities \cite{guarnera2020deepfake}. CNN architectures, notably XceptionNet \cite{faceforensics} enhanced with attention \cite{nguyen2019multi}, have significantly improved artifact detection. To alleviate data scarcity and adapt to evolving deepfakes, self-supervised \cite{hsu2020deep} and semi-supervised methods \cite{cozzolino2018forensictransfer} have been proposed.

The emergence of diffusion models \cite{stable_diffusion} has introduced new challenges, as these methods often lack the grid-like artifacts typical of GANs \cite{nichol2021glide,ramesh2022hierarchical,saharia2022photorealistic}. Investigations into intrinsic local dimensionality \cite{lorenz2023detecting} and reconstruction errors \cite{wang2020cnn} have been undertaken, while advanced architectures such as CNNs and Transformers \cite{coco_fake} further push detection limits. Deep learning–based approaches \cite{marra2019incremental,goebel2020detection,wang2020cnn,liu2020global,hulzebosch2020detecting,guo2022robust,pu2022learning,hu2021improving,zhang2020face,liang2022exploring,Yan_2023_ICCV,zheng2024few,chen2024masked,lin2024detecting,fan2023synthesizing,zhang2023x,yang2023improving,fan2023attacking,chen2023harnessing} have achieved notable performance gains despite persistent fairness issues \cite{trinh2021examination,xu2022comprehensive,news2,nadimpalli2022gbdf,masood2022deepfakes}. Recent efforts address these challenges by disentangling demographic and forgery features \cite{Lin_2024_CVPR}, refining optimization strategies \cite{Ju_2024_WACV}, reexamining up-sampling effects, and augmenting latent spaces \cite{Yan_2024_CVPR}. Efficiency for real-time deployment is also underscored \cite{lanzino2024faster}.
Some studies have aimed to understand and improve deepfake generalization. For instance, \cite{10377057} reframes the generalization problem from a game-theoretical perspective, \cite{li2023generalizable} examines the zero-shot performance of detectors and the contribution of individual neurons to generalization, and \cite{yan2024generalizing} investigates how video blending influences the generalization of video deepfake detectors.
 \\ 
A promising trend is integrating CL into DFD systems. \cite{usmani2024enhancing} presents an approach that improves robustness by updating the model incrementally without needing full retraining. Specifically, \cite{li2023continual} introduces a CL benchmark for DFD that presents three evaluation sequences based on difficulty or length and applies various CL methods in CIL and DIL settings. Building on this, \cite{TASSONE2024104143} enhances the benchmark by incorporating a new generator and evaluates two CL methods on it, while \cite{ieee10034569} reframes deepfake detection in a CIL setup, focusing on generalization.
Despite advances in CL for DFD, key challenges remain. Notably, current methods ignore the chronological nature of real-world deployment, use limited datasets, and overlook practical constraints like computational efficiency. Contrarily, our approach simulates deepfake evolution over extended periods, enabling realistic evaluation of model adaptability, retention, and generalization for effective detection while preserving efficiency and real-time inference.

%% file: sec/4_method.tex
\section{Method}
\label{sec:method}
\subsection{Problem Formulation}

Traditional binary DFD systems assume simultaneous access to a complete dataset \(D\) composed of samples from a set of deepfake generators \(\mathcal{G} = \{g_1, g_2, \ldots, g_n\}\). The classifier \(C\) is trained by minimizing the empirical risk:
\begin{equation}
    \min_{C} \; \frac{1}{|D|} \sum_{(x, y) \in D} \ell(C(x), y)
\end{equation}
where each \((x, y)\) represents an input-label pair (with labels 0 for real and 1 for fake) and \(\ell\) is a suitable loss function, such as cross-entropy. When a new deepfake generator \(g_{n+1}\) emerges, the updated dataset becomes \(D' = D \cup D_{n+1}\), requiring a complete retraining of \(C\). This approach is computationally expensive and inefficient, especially as the number of generators increases.

To alleviate this problem, as previous works \cite{pan2023dfil}, we reformulate the problem within a DIL framework. In this setting, the classifier is updated sequentially: At time step \(t\), the classifier \(C_t\) is trained only on the current dataset \(D_t\) generated by \(g_t\), while still maintaining performance in all domains previously encountered. Specifically, \(t\) represents different stages of the simulation; each time step corresponds to a time span that can be adjusted as a hyperparameter. The optimization objective is given by:
\begin{equation}
    \min_{C_t} \; L(C_t, D_t) + \lambda \, \Omega(C_t, C_{t-1})
\end{equation}
where \(L(C_t, D_t)\) is the classification loss on the current domain, \(\Omega(C_t, C_{t-1})\) is a regularization term that enforces the retention of previously learned knowledge (thereby mitigating catastrophic forgetting), and \(\lambda\) is a hyperparameter that balances the trade-off between learning new information and preserving past information.
In our work, we further impose a chronological constraint on the DIL framework. That is, the sequence of deepfake generators—and consequently their corresponding datasets—are released and processed in their natural temporal order. Reordering based on similarity, or any reordering in general, is not allowed. Each dataset \(D_t\) is observed strictly in the order \(t = 1, 2, \ldots, T\), reflecting the evolving nature of deepfake generation techniques over time. Although the underlying formulation remains identical to standard DIL, this chronological constraint ensures that the classifier's adaptation is aligned with real-world scenarios where new generators are introduced sequentially.

\subsection{Binary Deepfake Detection under Chronologically Domain Incremental Learning}

In our approach, we enforce a strict temporal ordering of data acquisition that mirrors the natural release of deepfake generators. Specifically, the model is updated sequentially following the authentic timeline without any reordering or artificial sampling of past data. At each chosen interval, new balanced batches are integrated into the learning process according to timeline protocol (see Sec.~\ref{timeline_protocol}). The temporal distribution and key attributes of the datasets are summarized in Table~\ref{tab:datasets}.

\subsubsection{Timeline Protocol}
\label{timeline_protocol}
The timeline protocol simulates the natural evolution of deepfake generation by associating each dataset \(D_1, D_2, \ldots, D_N\) with its unique release date \(t_i\) (where \(t_1 < t_2 < \cdots < t_N\)). At each update interval, as illustrated in Fig.~\ref{fig:framework}, the selection of datasets is governed by an exponentially weighted random selection model, which mirrors the historical distribution of deepfake instances. Specifically, the probability of selecting dataset \(D_i\) is defined as:
\begin{equation}
    P(D_i) = \frac{0.5 \cdot (0.5^i)}{\sum_{j=0}^{N-1} 0.5 \cdot (0.5^j)}
\end{equation}
with the normalization condition:
\begin{equation}
    \sum_{i=0}^{N-1} P(D_i) = 1
\end{equation}
Datasets are sampled in reverse chronological order relative to the current simulation time, thereby prioritizing more recent data.
A crucial aspect of this simulation is its attempt to replicate the actual distribution of deepfake samples as observed at a given historical moment. In practice, at any specific point in time, adapting the model to new tasks would require sampling random deepfake instances from contemporary sources, such as social networks. This approach mirrors real-world deployment scenarios where a model is continuously updated to detect emerging deepfake techniques by leveraging naturally occurring data streams.
\subsection{Evaluation Metrics}
To quantify performance in the presented environment, we propose two metrics: \textit{Continual AUC (C-AUC)} and \textit{Forward Transfer AUC (FWT-AUC)}. These extend established CL concepts to address the unique challenges of DFD, where datasets are typically highly imbalanced. Conventional metrics, like accuracy, are often inadequate for such tasks as they assume a balanced class distribution, which does not reflect the nature of most real-world DFD datasets. The evolving generation techniques and significant class imbalance in deep fake datasets demand robustness beyond traditional accuracy metrics.

\subsubsection{Continual AUC}
C-AUC measures retention on previously encountered data:
\begin{equation}
    \text{C-AUC}(t) = \frac{1}{|D_{1:t}|} \sum_{i=1}^{|D_{1:t}|} [AUC(C_t, D_i)]
    \label{eq_cauc}
\end{equation}
where $D_{1:t}$ are the datasets presented up to time $t$ and $C_t$ is the detector at $t$.
This metric is related to \textit{Average Accuracy} and \textit{Backward Transfer} \cite{lopez2017gradient} but uses Area Under The Curve (AUC) instead of accuracy, better-capturing robustness to class imbalance. Measures \textbf{stability}\cite{Grossberg1982StudiesOM}, which refers to retention of knowledge in previous and current tasks.
\subsubsection{Forward Transfer AUC}
FWT-AUC quantifies generalization to future generators:
\begin{equation}
    \text{FWT-AUC}(t) = \frac{1}{|D_{t+1:}|} \sum_{i=1}^{|D_{t+1:}|} [AUC(C_t, D_i)]
    \label{eq_fwtauc}
\end{equation}
where $D_{t+1:}$ are the dataset not yet released a time t.
This metric aligns with \textit{Forward Transfer (FWT)} \cite{lopez2017gradient} but uses AUC to  evaluate generalization to future tasks.

%% file: sec/5_experiments.tex
\begin{table*}[ht!]
\centering
\renewcommand{\arraystretch}{1.2} 
\begin{minipage}{0.95\textwidth}
    \centering
    \resizebox{\textwidth}{!}{%
    \begin{tabular}{lcccccc}
    \toprule
    \textbf{Dataset} & \textbf{Type} & \textbf{Resolution} & \textbf{Data Count} & \textbf{Sources} & \textbf{Month} & \textbf{Year} \\
    \midrule
    DeepfakeTIMIT \cite{korshunov2018deepfakes}    & Video  & $128\times128$ & 620 videos  & Faceswap-GAN (based on VidTIMIT)  & July      & 2018 \\
    WildDeepfake \cite{zi2020wilddeepfake}     & Video  & Variable       & 7,314 face sequences (from 707 videos) & Internet      & October   & 2020 \\
    DFFD \cite{dang2020detection}              & Image  & Variable       & Heterogeneous content                & FFHQ, CelebA, FF++ & N/A    & 2020 \\
    FakeAVCelebv2 \cite{khalid2021fakeavceleb}   & Video  & Variable       & 20,000 face sequences                & Faceswap, FSGAN    & N/A    & 2021 \\
    COCOFake \cite{amoroso2023parents}          & Image  & Variable       & 1,200,000 images                     & Stable Diffusion v1.4, v2.0 & N/A & 2023
    \\

    CIFAKE \cite{bird2024cifake}            & Image  & Variable       & 120,000 images (60k real + 60k synthetic) & Stable Diffusion v1.4 & February & 2024  \\
    \bottomrule
    \end{tabular}%
    }
    \caption{Datasets used for the CL setup, including resolution, data count, sources, release month, and publication year.}
    \label{tab:datasets}
\end{minipage}
\end{table*}

\begin{table*}[ht]
\centering
\small
\begin{adjustbox}{max width=\textwidth}
\begin{tabular}{llccc|ccc|ccc}
\toprule
\multirow{2}{*}{\textbf{Strategy}} & \multirow{2}{*}{\textbf{Model}} & \multicolumn{3}{c|}{\textbf{Monthly Batches = 10}} & \multicolumn{3}{c|}{\textbf{Monthly Batches = 20}} & \multicolumn{3}{c}{\textbf{Monthly Batches = 50}} \\
\cmidrule(lr){3-5} \cmidrule(lr){6-8} \cmidrule(lr){9-11}
 & & AUC & Mean C-AUC & Mean FWT-AUC & AUC & Mean C-AUC & Mean FWT-AUC & AUC & Mean C-AUC & Mean FWT-AUC \\
\midrule
\textbf{Full Retraining}          & \multirow{9}{*}{MobileNetV4-M \cite{qin2024mobilenetv4}} & 0.871 & -     & -     & 0.8711 & -     & -     & 0.871 & -     & - \\
\midrule
CLS-ER+EWC                      &  & \textbf{0.765}  & \textbf{0.803} & 0.520 & \textbf{0.793}  & 0.801 & 0.545 & \textbf{0.852}  & \textbf{0.857} & 0.482 \\
Replay+EWC                      &  & 0.745  & 0.792 & 0.527 & 0.760  & \textbf{0.822} & 0.469 & 0.841  & 0.840 & 0.504 \\
EWC \cite{kirkpatrick2017overcoming} &  & 0.733  & 0.784 & 0.475 & 0.743  & 0.815 & 0.503 & 0.806  & 0.753 & 0.499 \\
ESMER\cite{sarfrazerror}         & & 0.727  & 0.739 & 0.519 & 0.759  & 0.776 & 0.573 & 0.741  & 0.664 & 0.528 \\
DER++\cite{buzzega2020dark}      &  & 0.720  & 0.798 & 0.547 & 0.662  & 0.785 & 0.554 & 0.746  & 0.778 & 0.518 \\
Replay\cite{rolnick2019experience}    &  & 0.681  & 0.758 & 0.509 & 0.711  & 0.691 & 0.528 & 0.740  & 0.752 & 0.521 \\
CLS-ER \cite{arani2022learning}  &  & 0.640  & 0.746 & 0.493 & 0.635  & 0.784 & 0.523 & 0.703  & 0.569 & 0.492 \\
Naive                           &  & 0.607  & 0.742 & 0.518 & 0.567  & 0.712 & 0.520 & 0.593  & 0.702 & 0.515 \\
\midrule
\textbf{Full Retraining}          & \multirow{9}{*}{ViT-Tiny \cite{dosovitskiy2020image}}     & 0.960 & - & - & 0.960 & - & - & 0.960 & - & - \\

\midrule CLS-ER+EWC                      &    & 0.673 & 0.718 & 0.556 & 0.768 & 0.818 & 0.507 & 0.796 & 0.807 & 0.518\\
Replay+EWC                      &    & 0.679 & 0.699 & 0.552 & 0.799 & 0.779 & 0.537 & 0.818 & 0.824 & 0.512 \\
EWC \cite{kirkpatrick2017overcoming} &    & 0.663 & 0.702 & 0.498 & 0.781 & 0.764 & 0.516 & 0.808 & 0.808 & 0.520 \\
ESMER\cite{sarfrazerror}         &    & 0.556 & 0.667 & 0.510 &  \textbf{0.839} & 0.813 & 0.490 & 0.783 & 0.857 & 0.504 \\
DER++\cite{buzzega2020dark}      &    & \textbf{0.704} & 0.724 & 0.47 & 0.826 & 0.789 & 0.504 & 0.808 & 0.811 & 0.517 \\
Replay\cite{rolnick2019experience}    &    & 0.637 & 0.709 & 0.563 & 0.790 & 0.777 & 0.515 & 0.798 & 0.813 & 0.530 \\
CLS-ER \cite{arani2022learning}  &    & 0.700 & \textbf{0.763} & 0.507 & 0.793 & \textbf{0.824} & 0.496 & \textbf{0.932} & \textbf{0.915} &	0.528 \\
Naive                           &    & 0.662 & 0.700 & 0.506 & 0.616 & 0.683 & 0.476 & 0.777 & 0.798 &	0.512 \\
\midrule
\textbf{Full Retraining}          & \multirow{9}{*}{ConvNeXtV2-ATTO \cite{woo2023convnext}}   & 0.861 & - & - & 0.861 & - & - & 0.861 & - & - \\
\midrule
CLS-ER+EWC                      &     & \textbf{0.719} & \textbf{0.804} & 0.502 & 0.706 & \textbf{0.785} & 0.539 & 0.681 & 0.722 & 0.527 \\
Replay+EWC                      &     & 0.638 & 0.700 & 0.537 & 0.748 & 0.776 & 0.545 & 0.758 & 0.776 & 0.526 \\
EWC \cite{kirkpatrick2017overcoming} &     & 0.664 & 0.722 & 0.535 & 0.705 & 0.746 & 0.530 & 0.744 & 0.751 & 0.540 \\
ESMER\cite{sarfrazerror}         &     & 0.631 & 0.713 & 0.522 & 0.548 & 0.704 & 0.554 & 0.811 & 0.821 & 0.505 \\
DER++\cite{buzzega2020dark}      &     & 0.689 & 0.742 & 0.532 & 0.728 & 0.764 & 0.551 & 0.764 &0.783 & 0.539 \\
Replay\cite{rolnick2019experience}    &     & 0.678 & 0.706 & 0.540 & \textbf{0.753} & 0.777 & 0.545 & 0.766 & 0.770 & 0.519 \\
CLS-ER \cite{arani2022learning}  &     & 0.706 & 0.746 & 0.547 & 0.748 & 0.767 & 0.530 &  \textbf{0.827} & \textbf{0.857} & 0.499 \\
Naive                           &     & 0.664 & 0.722 & 0.535 & 0.492 & 0.637 & 0.473 & 0.636 & 0.720 & 0.541 \\
\midrule
\textbf{Full Retraining}          & \multirow{9}{*}{ FastViT-T8 \cite{vasu2023fastvit}}    & 0.726 & - & - & 0.726 & - & - & 0.726 & - & - \\
\midrule
CLS-ER+EWC                      &     & \textbf{0.686} & 0.651 & 0.502 & 0.584 & 0.642 & 0.530 & 0.587 & 0.613 & 0.491 \\
Replay+EWC                      &     & 0.468 & \textbf{0.699} & 0.518 & 0.452 & 0.663 & 0.503 & 0.636 & 0.643 & 0.502 \\
EWC \cite{kirkpatrick2017overcoming} &     & 0.568 & 0.669 & 0.506 & 0.637 & 0.632 & 0.505 & 0.477 & 0.704 & 0.520 \\
ESMER\cite{sarfrazerror}         &     & 0.634 & 0.651 & 0.527 & \textbf{0.663} & \textbf{0.689} & 0.571 & 0.624 & 0.611 & 0.511 \\
DER++\cite{buzzega2020dark}      &     & 0.573 & 0.666 & 0.527 & 0.600 & 0.659 & 0.557 & 0.573 & 0.556 & 0.542 \\
Replay\cite{rolnick2019experience}    &     & 0.593 & 0.555 & 0.527 & 0.628 & 0.631 & 0.559 & 0.558 & \textbf{0.752} & 0.513 \\
CLS-ER \cite{arani2022learning}  &     & 0.478 & 0.671 & 0.551 & 0.459 & 0.669 & 0.528 & \textbf{0.705} & 0.730 & 0.536 \\
Naive                           &     & 0.529 & 0.647 & 0.543 & 0.452 & 0.459 & 0.509 & 0.502 & 0.696 & 0.530 \\
\bottomrule
\end{tabular}
\end{adjustbox}
\caption{Evaluation of strategies across different monthly batches on the evaluation set of the datasets for four different models.}
\label{tab:experiments_models}
\end{table*}
\begin{table*}[ht]

\centering
\small
\begin{adjustbox}{max width=\textwidth}
\renewcommand{\arraystretch}{0.85}
\begin{tabular}{llcccc|cccc|cccc}
\toprule
\multirow{2}{*}{\textbf{Method}} & \multirow{2}{*}{\textbf{Model}} & \multicolumn{4}{c|}{\textbf{Monthly Batches = 10}} & \multicolumn{4}{c|}{\textbf{Monthly Batches = 20}} & \multicolumn{4}{c}{\textbf{Monthly Batches = 50}} \\
\cmidrule(lr){3-6} \cmidrule(lr){7-10} \cmidrule(lr){11-14}
 &  & Monthly GPU Time (s) & Total GPU Time (min) & Unique Samples &  & Monthly GPU Time (s) & Total GPU Time (min) & Unique Samples &  & Monthly GPU Time (s) & Total GPU Time (min) & Unique Samples &  \\
\midrule
\textbf{Full Retraining} & \multirow{9}{*}{MobileNetV4-M \cite{qin2024mobilenetv4}} & - & 3130.02 * \( \mathcal{N} \) & 5773248 &  & - & 3130.02 * \( \mathcal{N} \) & 5773248 &  & - & 3130.02 * \( \mathcal{N} \) & 5773248 &  \\
\midrule
CLS-ER+EWC &  & 3.5 & 4.67 &\multirow{8}{*}{12800}  &  & 4.73 & 6.31 & \multirow{8}{*}{25600} &  & 13.76 & 18.35 &\multirow{8}{*}{64000}  &  \\
Replay+EWC &  & 3.11 & 4.15 &  &  & 4.25 & 5.67 &  &  & 10.09 & 13.45 &  &  \\
EWC \cite{kirkpatrick2017overcoming} &  & 0.91 & 1.21 &  &  & 1.75 & 2.33 &  &  & 4.35 & 5.80 &  &  \\
ESMER\cite{sarfrazerror} &  & 2.25 & 3.00 & & & 4.66 & 6.21 &  &  & 11.38 & 15.17 &  &  \\
DER++\cite{buzzega2020dark} &  & 1.69 & 2.25 & &  & 3.08 & 4.11 &  &  & 8.06 & 10.75 &  &  \\
Replay \cite{rolnick2019experience} &  & 2.54 & 3.39 &   &  & 3.24 & 4.32 &  &  & 8.04 & 10.72 &  &  \\
CLS-ER \cite{arani2022learning} &  & 3.25 & 4.33 & &  & 4.47 & 5.96 &  &  & 12.41 & 16.55 &  &  \\
Naive & & 0.9 & 1.20 &  &  & 0.99 & 1.32 &  &  & 2.5 & 3.33 &  &  \\
\midrule
\textbf{Full Retraining} & \multirow{9}{*}{ViT-Tiny \cite{dosovitskiy2020image}} & - & 4451.55 * \( \mathcal{N} \) & 5773248 &  & - & 4451.55 * \( \mathcal{N} \) & 5773248 &  & - & 4451.55 * \( \mathcal{N} \) & 5773248 &  \\
\midrule
CLS-ER+EWC &  & 1.52 & 2.03  & \multirow{8}{*}{12800} & & 2.39 & 3.19  & \multirow{8}{*}{25600}  & & 9.52 & 12.69 &\multirow{8}{*}{64000}  &\\
Replay+EWC &  & 1.32 & 1.76  & & & 1.16 & 1.55  & & & 2.79 & 3.73 &  &  \\
EWC \cite{kirkpatrick2017overcoming} &  & 0.48 & 0.64  & & & 0.92 & 1.23  & & & 2.46 & 3.27 &  &  \\
ESMER\cite{sarfrazerror} &  & 1.19 & 1.59  & & & 3.09 & 4.13  & & & 10.26 & 13.68 &  &  \\
DER++\cite{buzzega2020dark} &  & 0.88 & 1.18  & & & 1.18 & 1.57  & & & 3.01 & 4.01 &  &  \\
Replay \cite{rolnick2019experience} &  & 0.49 & 0.65  & & & 0.77 & 1.02  & & & 3.53 & 4.71 &  &  \\
CLS-ER \cite{arani2022learning} &  & 0.28 & 0.38  & & & 3.61 & 4.81  & & & 5.11 & 6.81 &  &  \\
Naive & & 1.28 & 1.71  & & & 0.53 & 0.70  & & & 1.45 & 1.93 &  &  \\

\midrule
\textbf{Full Retraining} & \multirow{9}{*}{ConvNeXtV2-ATTO \cite{woo2023convnext}} & - & 930 * \( \mathcal{N} \) & 5773248 &  & - & 930 * \( \mathcal{N} \) & 5773248 &  & - & 930 * \( \mathcal{N} \) & 5773248 &  \\
\midrule
CLS-ER+EWC &  & 1.53 & 2.04 & \multirow{8}{*}{12800} & & 2.37 & 3.17 & \multirow{8}{*}{25600} & & 5.72 & 7.63 & \multirow{8}{*}{64000} & \\
Replay+EWC &  & 0.64 & 0.85 & & & 1.14 & 1.53 & & & 2.58 & 3.44 & & \\
EWC \cite{kirkpatrick2017overcoming} &  & 0.47 & 0.63 & & & 0.76 & 1.01 & & & 2.20 & 2.94 & & \\
ESMER\cite{sarfrazerror} &  & 1.76 & 2.35 & & & 2.41 & 3.22 & & & 7.53 & 10.04 & & \\
DER++\cite{buzzega2020dark} &  & 0.97 & 1.29 & & & 1.33 & 1.77 & & & 3.37 & 4.50 & & \\
Replay \cite{rolnick2019experience} &  & 0.49 & 0.66 & & & 0.72 & 0.96 & & & 1.89 & 2.52 & & \\
CLS-ER \cite{arani2022learning} &  & 1.42 & 1.89 & & & 2.09 & 2.79 & & & 5.12 & 6.83 & & \\
Naive & & 0.27 & 0.36 & & & 0.50 & 0.66 & & & 1.39 & 1.85 & & \\

\midrule
\textbf{Full Retraining} & \multirow{9}{*}{FastViT-T8 \cite{vasu2023fastvit}} & - & 1.669.33 * \( \mathcal{N} \) & 5773248 &  & - & 1.669.33 * \( \mathcal{N} \) & 5773248 &  & - & 1.669.33 * \( \mathcal{N} \) & 5773248 &  \\
\midrule
CLS-ER+EWC &  & 2.59 & 3.46 & \multirow{8}{*}{12800} & & 4.27 & 5.70 & \multirow{8}{*}{25600} & & 9.91 & 13.22 & \multirow{8}{*}{64000} & \\
Replay+EWC &  & 2.33 & 3.11 & & & 1.92 & 2.56 & & & 4.68 & 6.24 & & \\
EWC \cite{kirkpatrick2017overcoming} &  & 0.83 & 1.10 & & & 1.41 & 1.89 & & & 4.01 & 5.35 & &  \\
ESMER\cite{sarfrazerror} &  & 3.30 & 4.40 & & & 5.45 & 7.26 & & & 9.55 & 12.74 & &  \\
DER++\cite{buzzega2020dark} &  & 1.72 & 2.30 & & & 1.94 & 2.58 & & & 4.94 & 6.59 & &  \\
Replay \cite{rolnick2019experience} &  & 0.96 & 1.28 & & & 1.30 & 1.74 & & & 3.01 & 4.02 & &  \\
CLS-ER \cite{arani2022learning} &  & 2.22 & 2.97 & & & 3.87 & 5.17 & & & 8.57 & 11.43 & &  \\
Naive & & 0.47 & 0.63 & & & 0.77 & 1.03 & & & 2.59 & 3.45 & &  \\
\bottomrule
\end{tabular}
\end{adjustbox}
\caption{Comparison of GPU time and unique samples seen across different monthly batches. "Monthly GPU Time (s)" represents the monthly processing time in seconds, "Total GPU Time (min)" indicates the total GPU time in minutes over the entire simulation, and "Unique Samples" refers to the number of distinct samples encountered. The Full Retraining remains independent of monthly batches. The variable \( \mathcal{N} \) denotes the number of adaptations training, where \( \mathcal{N} = 1 \) represents a single training session, \( \mathcal{N} = 2 \) indicates one retraining iteration, and so on, ensuring adaptation to newly emerging generators.}
\label{tab:gpu}
\end{table*}

\begin{figure*}[ht]
    \centering
    \begin{tabular}{cccc}
        \adjustbox{trim=0.4cm 0cm 0.0cm 0cm,clip}{
            \includegraphics[width=0.25\textwidth]{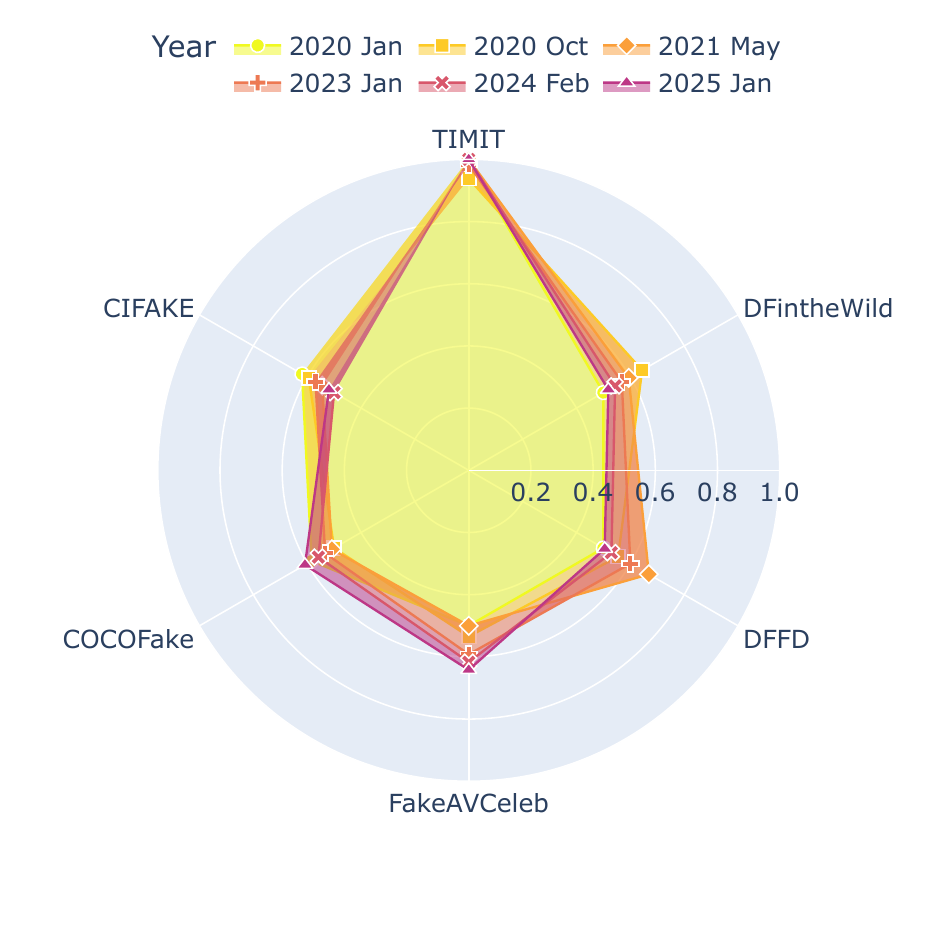}} & 
        \adjustbox{trim=0.4cm 0cm 0.0cm 0cm,clip}{
            \includegraphics[width=0.25\textwidth]{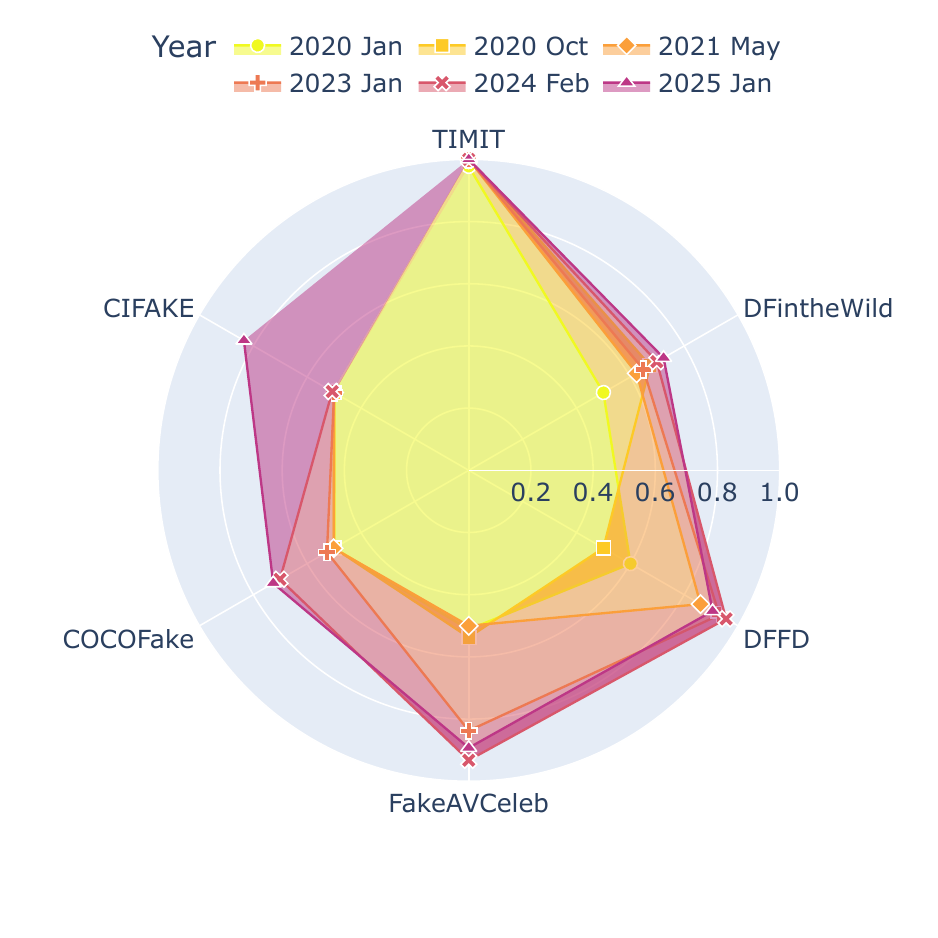}} &  
        \adjustbox{trim=0.4cm 0cm 0.0cm 0cm,clip}{
            \includegraphics[width=0.25\textwidth]{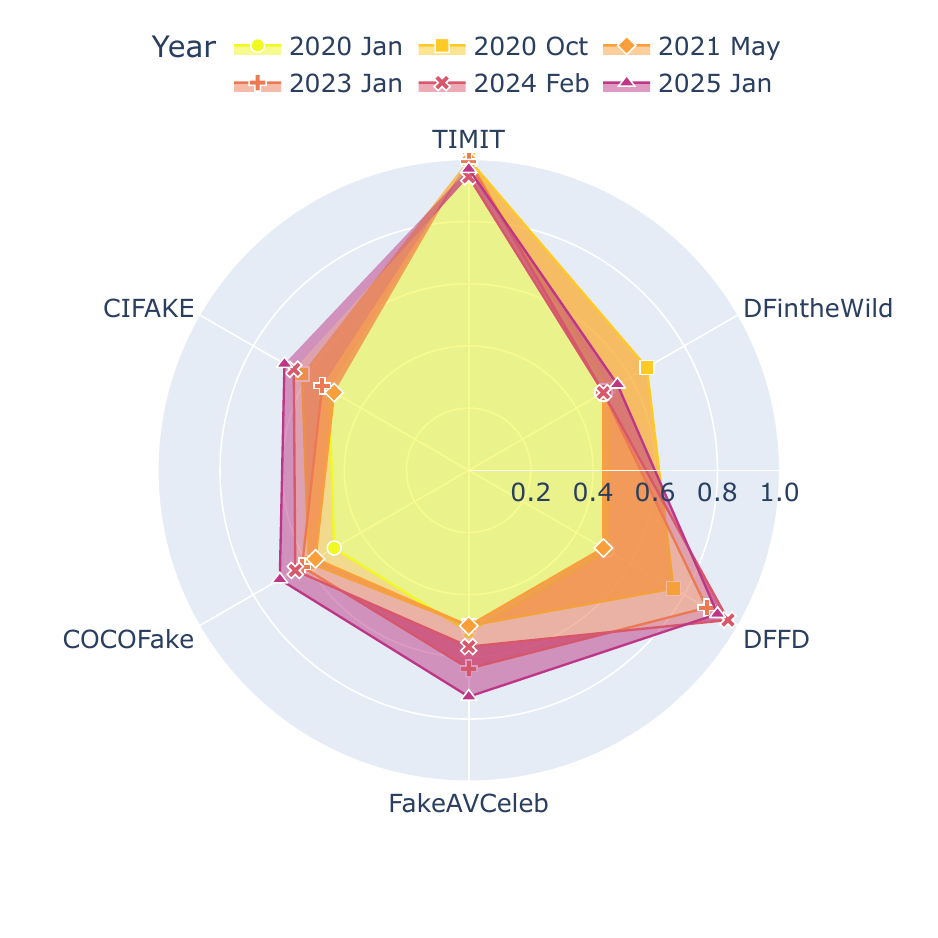}} &  
        \adjustbox{trim=0.4cm 0cm 0.0cm 0cm,clip}{
            \includegraphics[width=0.24\textwidth]{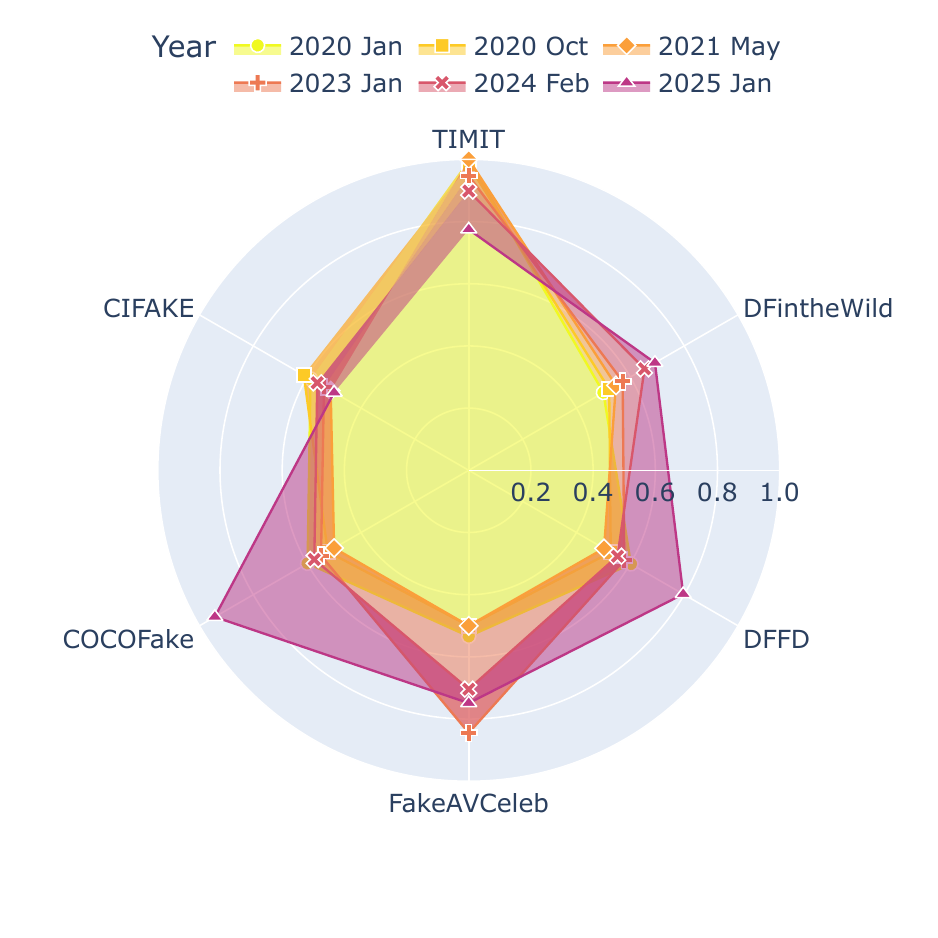}} \\  
        \small (a) Naive Strategy & \small (b) CLS-ER+EWC Strategy & \small (c) ESMER Strategy & \small (d) Replay Strategy
    \end{tabular}
    \caption{Spider charts showing the evaluation AUC on each dataset for each evaluation step using different strategies.}
    \label{fig:spider_charts}
\end{figure*}
\section{Experiments}
In the first part of this section, we clarify the choice of models and the CL strategies that were tested, and then we present the results.
\subsection{Models}
To balance efficiency in both training and inference, we employ a subset of lightweight models, specifically: FastViT-T8 \cite{vasu2023fastvit}, ConvNeXtV2-ATTO \cite{woo2023convnext}, ViT-Tiny \cite{dosovitskiy2020image}, and MobileNetV4-Medium \cite{qin2024mobilenetv4}. The size of these models was chosen based on the GPU time required for single-task training, ensuring a balance between efficiency and performance. We set the input size to 384x384.
Our method utilizes a single, compact model that relies solely on visual data, excluding the vocal aspects of deepfakes. Each video sample consists of five consecutive frames, allowing the model to effectively capture temporal patterns and visual cues. The frames are processed individually, and the final class prediction is obtained by averaging the highest probabilities across frames.
\subsection{Continual learning Strategies} A diverse set of CL strategies was used. The \textbf{Naive} approach follows a simple sequential training paradigm without integrating any continual learning mechanisms. In contrast, \textbf{Replay} \cite{rolnick2019experience} mitigates forgetting by storing and reusing samples from previous tasks. Regularization-based techniques such as \textbf{EWC} \cite{kirkpatrick2017overcoming} (Elastic Weight Consolidation) impose penalties on changes to parameters identified as crucial for past tasks. Extending replay methods, \textbf{CLS-ER} \cite{arani2022learning} (Class-Semantic Replay) refines memory replay by incorporating class-semantic information, while \textbf{CLS-ER+EWC} combines replay with regularization for improved retention. A more adaptive replay mechanism, \textbf{ESMER} \cite{sarfrazerror} (Error-Sensitive Memory Replay), adjusts sample contributions according to their error sensitivity. Furthermore, \textbf{DER++} \cite{buzzega2020dark} (Dark Experience Replay++) enhances replay strategies through knowledge distillation. Lastly, a hybrid approach, \textbf{Replay+EWC}, synergistically integrates the strengths of both Replay and EWC. Each of these methods presents distinct advantages and trade-offs in combating catastrophic forgetting, and their comparative evaluation investigates effective CL methodologies.
\subsection{Full Retraining Baselines}
To establish a baseline model that serves as a reference for all continual learning strategies, we fine-tuned the models pre-trained on ImageNet21K \cite{ridnik2021imagenet} using the datasets presented in Tab. \ref{tab:datasets}. The training was performed using the AdamW optimizer with a learning rate of \(10^{-6}\) and employed a Cosine Annealing Learning Rate Scheduler with \(T_{\text{max}} = 100,000\) iterations. Each batch comprised 16 samples per dataset (resulting in a total batch size of 96 samples) with a balanced composition of 8 real and 8 fake samples per subset, and training was carried out for 100,000 iterations. The baseline models were subsequently evaluated on the test sections of the six datasets mentioned above.
\subsection{Experimental Setup}
All experiments were conducted on a 14th-generation Intel i9 processor and an NVIDIA RTX 4090 GPU. We evaluate each CL method within the presented framework. The tasks are introduced on a monthly basis. For every newly released dataset, a complete model evaluation is performed, including the AUC, C-AUC, FWT-AUC, and GPU time. The experiments are categorized into three groups based on the number of monthly tasks: $10$, $20$, and $50$. For each scenario, the models are trained with the corresponding number of batches per simulated month. Consequently, during the simulation, each method processes a total of $80 \times \text{monthly batches} \times 16$ samples, where $16$ represents the batch size. AdamW is used as the optimizer for all experiments.
For the ESMER method, we set the parameters as follows: \(\beta = 1\), \(\alpha = 0.9\), and the update decay for the stable model weight is set to $0.99$. For EWC and Replay+EWC, the stable model weight update decay is set to $0.99$. The CLS-ER and CLS-ER+EWC use a plastic model update decay of $0.99$ and a stable model update decay of $0.999$.
When the monthly batch size is $50$, the learning rate for the Naive, Replay, CLS-ER+EWC, Replay+EWC, EWC, and DER++ versions is set to \(1 \times 10^{-5}\), while for ESMER and CLS-ER, the learning rate is set to \(1 \times 10^{-4}\). The memory buffer size is set to $50$ for all techniques except for Replay, which has a memory buffer size of $10$. Additionally, for Replay, Replay+EWC, and CLS-ER, the regularization coefficient \(\lambda\) is set to $10$. For ESMER and CLS-ER+EWC, \(\lambda = 0.5\), and for EWC, \(\lambda = 0.1\).
When the monthly batch size is $20$, the learning rate for Naive is set to \(1 \times 10^{-3}\), and for Replay, CLS-ER+EWC, Replay+EWC, EWC, and DER++, the learning rate is \(1 \times 10^{-5}\). For CLS-ER and ESMER, the learning rate is set to \(1 \times 10^{-4}\). The memory buffer size is $100$ for CLS-ER+EWC, ESMER, and CLS-ER, and $50$ for the other techniques. For Replay+EWC, CLS-ER+EWC, and CLS-ER, the regularization coefficient \(\lambda\) is set to $10$. For Replay, \(\lambda\) is set to $1$, and for DER++, \(\lambda\) is set to $0.5$.
When the monthly batch size is $10$, the learning rate for Naive, DER++, and EWC is set to \(1 \times 10^{-5}\), while for CLS-ER+EWC, Replay+EWC, ESMER, Replay, and CLS-ER, the learning rate is set to \(1 \times 10^{-4}\). The memory buffer size is $100$ for CLS-ER+EWC, Replay+EWC, ESMER, DER++, Replay, and CLS-ER.
\subsection{Results}
\label{subsec:results}
\noindent\textbf{Performance of CL Strategies:}
Tab. \ref{tab:experiments_models} compares the performance of CL strategies with the complete baseline of retraining. The results indicate that all CL approaches outperform the naive method. Moreover, in the largest monthly batch setting, the methods match the rooftops closely.
CLS-ER+EWC and Replay+EWC, which incorporate both replay and semantic memory mechanisms, perform well in the monthly batch 10 configuration. However, in the monthly batch 20 and 50 setups, each model favors a specific approach, suggesting that each method would benefit from its own dedicated hyperparameter tuning.
The best overall performance is achieved with ViT-T in the monthly batch 50 setup, reaching an AUC of 0.932 and a Mean C-AUC of 0.915. FastViT exhibits the lowest performance.
Overall, increasing monthly batches improves C-AUC, highlighting the importance of sufficient data for stability. In contrast, regularization-only methods like EWC show limited scalability.
\noindent\textbf{Generalization Limitations:}
All strategies exhibit near-random FWT-AUC ($0.49-0.57$), regardless of monthly batch configuration, method or model. Even ViT-T with CLS-ER, the best-performing strategy is random guessing, confirming that each generator introduces unique artifacts.
\noindent\textbf{Computational Efficiency:}
Tab. \ref{tab:gpu} reveals a clear efficiency hierarchy. Naive fine-tuning requires minimal overhead (0.27-2.59 min total GPU time), while combined replay-regularization methods (CLS-ER+EWC) demand 0.64-18.35 min. Despite this overhead, all continual methods remain orders of magnitude more efficient than full retraining ($155$ times less GPU Time). The chosen architectures enable real-time inference, making even intensive strategies viable on mobile devices.
\noindent\textbf{Tradeoffs Between Stability and Generalization:}
While C-AUC improves with an increased number of monthly batches, FWT-AUC remains relatively stagnant. For instance, Replay+EWC on MobileNetV4 boosts C-AUC by 7.8\% (from 0.792 to 0.840) when scaling from 10 to 50 batches, yet its FWT-AUC only fluctuates between 0.469 and 0.504. This suggests that stability, as measured by C-AUC, is influenced by the replay buffer size, the regularization strength, method and model, whereas generalization, indicated by FWT-AUC, is fundamentally constrained by distribution shifts between generators. The absence of a correlation between C-AUC and FWT-AUC confirms that these objectives are distinct. Although replay mechanisms help mitigate forgetting, they do not facilitate transfer learning across non-stationary generator distributions.
\subsection{The Non-Universal Deepfake Distribution Hypothesis}  
\label{ssec:transfer_hypothesis}
In all the conducted experiments, despite extensive hyperparameter tuning, different methods and models, the \textit{FWT-AUC} consistently remains close to $0.5$. This result indicates the classifier behaves randomly on future generators. This suggests that each generator leaves a unique signature, and no "distribution of deepfakes" can be generalized.
Moreover, this highlights that we cannot rely on a "natural distribution" of deep-fake media. Not all edited images or videos qualify as deepfakes, as naive modifications—such as simple color adjustments, font changes, or montages—do not constitute deepfake manipulations.
\\
The \emph{Non-Universal Deepfake Distribution Hypothesis} seeks to capture the temporal evolution of a detector’s effectiveness as the underlying deepfake generators evolve over time.
The hypothesis highlights the need for CL that can respond to the rapid evolution of deepfake generators, ensuring sustained robustness in real-world applications.\\

\textbf{Hypothesis Statement:} Deepfake detection cannot be generalized through static training.
\subsubsection{Formulation}
Let $\mathcal{G}$ denote the space of deepfake generators evolving through discrete time steps.  For any temporal window $T=[1,N]\subseteq \mathbb{N} \text{ with } t \in T$, let $C_{1:t}^s$ represent a classifier trained on synthetic images of generators ${G}_{1:t}$ using the strategy $s \in \mathcal{S}$, where $\mathcal{S}$ is the set of training methodology and hyperparameter configuration.
Let $ \mathcal{D}(G)$ be the function that returns a dataset comprising real and fake images, given a generator or a set of generators.
\noindent\textbf{Transferability Semantics:} The \emph{maximum temporal transferability} quantifies the classifier’s ability to generalize from past data to the next generator:
\begin{equation}
\mathcal{T}_{\max} = \max_{\substack{ t \in T \text{, }s \in \mathcal{S}}} \left( \underbrace{\text{AUC}\left(C_{1:t}^s, \mathcal{D}(G_{t+1})\right)}_{\text{Future generalization}} - 0.5 \right)
\label{eq:max_transfer}
\end{equation}
\textbf{Note:} We subtract $0.5$ because an AUC of $0.5$ corresponds to the performance of random guessing in binary classification. This normalization centers the metric at zero, making it easier to interpret.

\noindent\textbf{Decay Dynamics:} The \emph{transfer decay factor} quantifies progressive performance erosion: \begin{equation}
\mathcal{T}_{\text{decay}} = \mathbb{E}_{s \in S,t \in T} \left[ \frac{\text{AUC}(C_{1:t}^s,\mathcal{D}(G_{t+2})) - 0.5}{\text{AUC}(C_{1:t}^s, \mathcal{D}(G_{t+1})) - 0.5} \right]
\end{equation}

In the case $\mathcal{T}_{\text{decay}} < 1$ it leads to a compounded degradation model for $k$-step generalization even for the best strategy:\begin{equation}
\mathcal{T}_{\text{comp}} = 0.5 + \mathcal{T}_{\max} \cdot (\mathcal{T}_{\text{decay}})^{k}
\label{eq:compounded_decay}
\end{equation}

\noindent\textbf{Empirical Parameterization:} Under experimental conditions, the temporal window \( T \approx \overline{T} = [1,6] \) aligns with the dataset release chronology presented in the method. The strategy space \( \mathcal{S} \approx \overline{\mathcal{S}} \) includes all CL methods used in the methods section and all hyperparameter configurations in the ablation study, with \( |\overline{\mathcal{S}}| = 618 \). Additionally, \(\mathcal{D}(G_k) \approx D_k\) represents the test set of the k-th dataset.

Using this setup we derive:
\begin{equation}
\mathcal{T}_{\max} \approx \overline{\mathcal{T}_{\max}} = \max_{\substack{ t \in \overline{T} \text{, }s \in \overline{\mathcal{S}}}} \left( \underbrace{\text{AUC}\left(C_{1:t}^s, D_{t+1}\right)}_{\text{Future generalization}} - 0.5 \right)
\end{equation}

The value \( \overline{\mathcal{T}_{\max}} = 0.094 \) is observed across all CL strategies and hyperparameters, highlighting fundamental limitations in the classifier trained on historical data and evaluated on data generated on the next generators. \begin{equation}
\mathcal{T}_{\text{decay}} \approx \overline{\mathcal{T}_{\text{decay}}} = \frac{1}{|\overline{S}||\overline{T}|} \sum_{s \in \overline{S}} \sum_{t \in \overline{T}} \frac{\text{AUC}(C_{1:t}^s, D_{t+2}) - 0.5}{\text{AUC}(C_{1:t}^s, D_{t+1}) - 0.5}
\end{equation}
The empirical value \( \overline{\mathcal{T}_{\text{decay}}} = 0.54 \), calculated across all CL strategies and hyperparameters (selecting only experiments where the classifier achieved at least 0.75 on evaluation AUC), indicates that residual classification capacity decays exponentially. This leads to a compounding degradation process:\begin{align}
\mathcal{T}_{\text{comp}} \approx  \overline{\mathcal{T}_{\text{comp}}}  &= 0.5 + (\overline{\mathcal{T}_{\max}}) \cdot (\overline{\mathcal{T}_{\text{decay}}})^{k} \nonumber = \\ 
&= 0.515 \approx \text{Random guessing}
\end{align}
This empirical characterization suggests that \emph{any static detector} \( C_{1:t}^s \) inevitably converges to random guessing against evolving generators \( G_{t+k} \) in a very short time in the setup proposed even the best strategy turns to random guessing in $3$ time units.

%% file: sec/6_ablation_study.tex
\subsection{Ablation Study}  
In our ablation study conducted on MobileNetV4, we evaluate several critical hyperparameters. We test three different learning rates—\(10^{-5}\), \(10^{-4}\), and \(10^{-3}\)—which determine the optimization step size and significantly affect the model's learning effectiveness without overfitting. The memory buffer size, which dictates the number of batches stored for replay-based strategies, is examined using two values: 50 and 100. Additionally, we assess the impact of the number of monthly batches by testing configurations with 10, 20, and 50 batches. For the methods EWC, Replay+EWC, ESMER, DER++, and CLS-ER+EWC, we explore various regularization strengths, setting \(\lambda_{\text{reg}}\) to 0.1, 0.5, 1, and 10, while for DER++ we also evaluate \(\alpha\) over the same range. 
Fig. \ref{fig:spider_charts} visualizes evaluation AUC trajectories of MobileNetV4 across datasets and time steps. CLS-ER+EWC (Fig. \ref{fig:spider_charts}b) maintains stable performance on historical datasets (C-AUC=$0.857$), while naive fine-tuning (\ref{fig:spider_charts}a) shows severe forgetting (C-AUC=$0.702$). All strategies exhibit radial symmetry in the spider charts, confirming the almost temporal independence of generator-specific detection performance. The evaluation spider charts for each configuration are provided in the supplementary material.

%% file: sec/7_conclusion.tex
\section{Conclusion}
\label{sec:conclusion}

This paper reframes DFD as a DIL problem with a chronological constraint, to simulate real-case scenario, addressing the challenge of adapting to evolving generation technologies while retaining knowledge of past techniques. Our framework, leveraging lightweight architectures and diverse CL strategies, demonstrates that CL enables efficient adaptation ($\sim$155 times less GPU time) in respect with full retraining while maintaining performance on previously encountered generators. However, the near-random FWT-AUC $\approx$ $0.5$ across all strategies empirically validates our \emph{Non-Universal Deepfake Distribution Hypothesis}: each generator leaves unique, non-transferable artifacts, rendering generalization to future techniques fundamentally limited, so DFD cannot be generalized through static training.

These findings reveal an exponential decay pattern in the detection capacity, where residual performance deteriorates to random guessing ($0.515$ AUC) in $3$ time steps. This underscores the inadequacy of static training paradigms and establishes the imperative for CL. Our results suggest that detection systems must prioritize two complementary objectives: (1) efficient retention of historical knowledge through replay and regularization mechanisms, and (2) rapid adaptation pipelines to incorporate emerging techniques with minimal effort.

Our framework demonstrates that deepfake detectors can be updated efficiently using a few data, enabling maintenance by independent actors with limited resources. This efficiency democratizes detection ecosystems, allowing researchers, small organizations, and even community collectives to train and update detectors without reliance on centralized infrastructure or massive computational budgets.